# Arabian Horse Identification Benchmark Dataset

Ayat Taha [1*], Ashraf Darwish [2*], and Aboul Ella Hassanien [3**]

[1] Faculty of Science, Alazhar University, Cairo, Egypt

[2] Faculty of Science, Helwan University, Cairo, Egypt

[3] Faculty of Computers and Information, Cairo University, Cairo, Egypt

Ayat_taha@ymail.com, ashraf.darwish.eg@ieee.org, aboitcairo@gmail.com

**Abstract**

The lack of a standard muzzle print database is a challenge for conducting researches in Arabian horse identification systems. Therefore, collecting a muzzle print images database is a crucial decision. The dataset presented in this paper is an option for the studies that need a dataset for testing and comparing the algorithms under development for Arabian horse identification. Our collected dataset consists of 300 color images that were collected from 50 Arabian horse muzzle species. This dataset has been collected from 50 Arabian horses with 6 muzzle print images each. A special care has been given to the quality of the collected images. The collected images cover different quality levels and degradation factors such as image rotation and image partiality for simulating real time identification operations. This dataset can be used to test the identification of Arabian horse system including the extracted features and the selected classifier.

## 1. Introduction

Despite the modernization of agriculture, the horse, in Arab countries holds a significant place in the rural socio-economic matrix. The Arabian horses are used in agriculture, light traction, riding and leisure activities, notably the fantasia. Animal identification has been widely performed in ecological research and in such fields as pedigree enrollment. Identifying animals by their external features (hair color and so on) or by marks (brands/ear tags) have been widely used methods. However, they are not ideal: for example, external feature identification is based on subjective judgment, and may be difficult in the case of horses without prominent features; and the application of marks such as brands causes the animals pain. For these reasons, identification technology research has begun to analyze the biometric features of animals by using computers [1]. The existing identification methods generally use external features such as hair color, white spots (patterns on head and legs), and hair whorls [2]. For this reason, identification must be quick and accurate, and its computerization is desirable so as to reduce labor and increase accuracy. Horses are sensitive animals and sometimes become skittish when touched by a foreign object. Therefore, muzzle, recognition, is useful for identifying racehorses.

---

*Member of the Scientific Research Group in Egypt (SRGE).

**Chairman of the Scientific Research Group in Egypt (SRGE).

Muzzle print, or nose print, was investigated as distinguished pattern for animals since 1921 [3]. It is considered as a unique animal identifier that is similar to human fingerprints. There are major problems in horse muzzle recognition are low image quality caused by movement of the horse during identification, muzzle variation resulting from environmental transitions, and the coarse fold pattern of the horse muzzle. Applying human biometric recognition technology to the horse muzzle raises three issues: (1) the quality of the acquired image data, (2) modeling the muzzle area, (3) the features of the folds to be used for recognition.

There is a lack of a standard muzzle print benchmark. Driven from this need, the first contribution of this research is to collect a database of live captured muzzle print images that works as a benchmark dataset for Arabian horse identification approach.

This paper presents a new data set of horse muzzle patterns, which leads to the possibility of horse identification. A test system incorporating the chosen dataset was constructed and its effectiveness was examined by recognition ability evaluations in the field.

Table1. Specifications about subject area of data set and it is application on computer science.

| Subject area | Computer Science. |
|---|---|
| More specification subject area | Image processing, Machine Learning, and animal Biometrics. |
| Type of data | Image. |
| How the data was acquired | Nokia Lumia 625 Camera |
| Experimental factors | Manual segmentation pre-processing applied. |
| subject area Data format | Jpeg image. |
| Experimental features | 300 Image with different image sizes. |
| Data source location | EL Zahraa Farm- Ain Shams- Egypt. |
| Data accessibility. | Data are available with this article. |

## 2. Materials and methods
### 2.1. Quality of acquired image data

In order to get a muzzle print image with good quality there are some steps must followed it accurately. First: to take the muzzle print image easily the Arabian horse should be in a stanchion, by holding its head and raises it slightly by someone and taking the muzzle image by someone else.   Second: it is necessary to wipe the nose before tokening image, Due to the fact that the Arabian horse perspires freely through the pores of the nose and due to the Remains of eating on the muzzle to avoid the noise on the image. Third: Using infrared illumination avoids making the horses nervous. The distance between the muzzle of the Arabian horse and the camera is approximately from 10 to 20 cm, and the focus and zoom are set so that the muzzle occupies more than three-quarters of the image. By applying the three steps we get muzzle print images with height quality which contains the grooves and valley of the muzzle with no noise.

Horses move their heads almost all the time even if their reins are being held. Therefore, it was very difficult to acquire numerous adequate images for processing. When we visually inspect the acquired images, only 10 to 20% are adequate for identification. It is important to exclude poor images in order to avoid recognition failure from the processing of poorly focused images or of images excluding the muzzle. Figure 1 shows different images of a horse muzzle.

The images contained the subcutaneous facial-nacial glands causing more or less pronounced elevations forming the irregular lines in form of grooves between these elevations. On some animals the elevations are less pronounced than on others, in which case the nose may be termed comparatively smooth. With "smooth" noses care must be taken must zoom the camera to focus on the grooves and valleys of the muzzle.

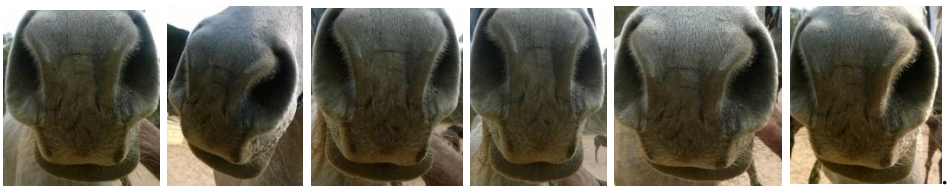

Fig. 1.Sample of six muzzle print of female Arabian horse and one image of its face.

### 2.2. Modeling the muzzle area

Animal muzzle prints have some discriminative features according to the grooves, or valleys, and beads structures. These uneven features are distributed over the skin surface in the Arabian horse nose area, and they are defined by the white skin grooves or by the black convenes surrounded by the grooves [4]. Return to Fig. 2 for consulting the convenes and the grooves in muzzle prints taken from two different male and female Arabian horses.

We believe that the rate of successful recognition of muzzle can drop because of the differences of muzzle condition or variations of image quality. Thus, no oval shapes should be used for muzzle modeling.

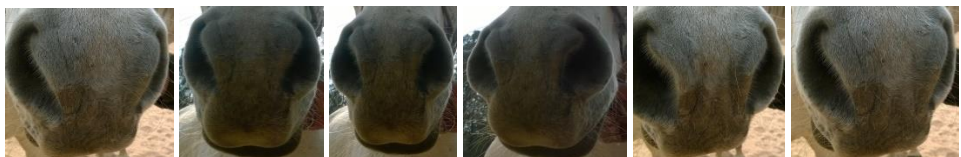

Fig. 2. Sample of six muzzle print of male Arabian horse and one image of its face.

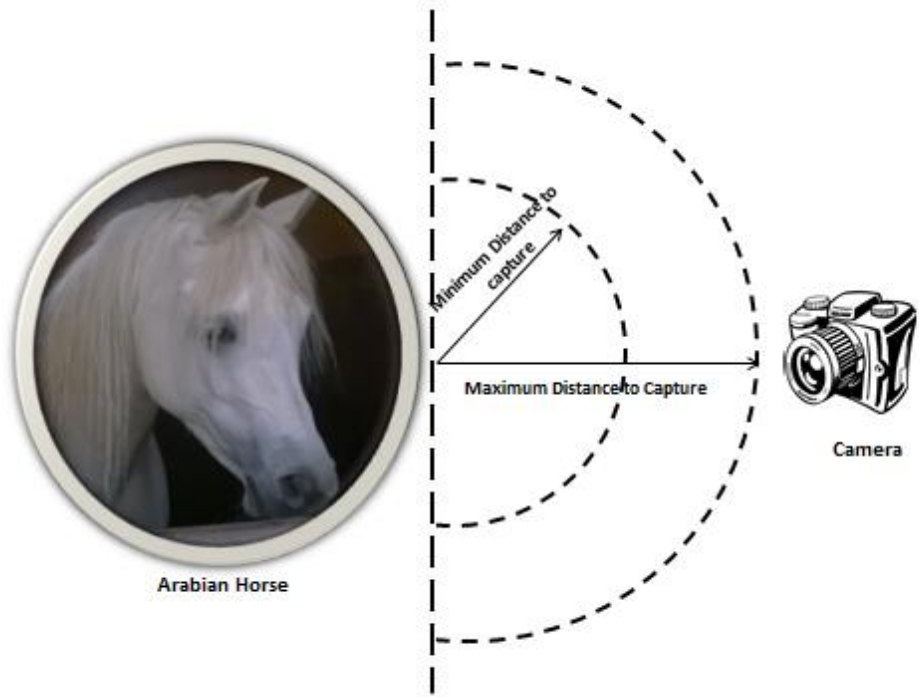

Fig.3. The specifications of the process of capturing muzzle print of Arabian horse

## 2.3. Fold features in recognition

The muzzle features of horses are not compact and they is complicated. This offers the possibility of highly accurate identification by an appropriate feature extraction method.

Table 2. Description of dataset classes and the dimensions of each class.

| Class | Number of Samples | Minimum Dimension (Pixel) | Maximum Dimension (Pixel) |
|---|---|---|---|
| Class 1 (Female) | 15× 6 | 600×700 | 1400×1600 |
| Class 2 (male) | 35× 6 | 1000×1200 | 1400×2100 |

## 2.4. Camera Description

Our dataset was collected by using a Nokia Lumia 625 digital camera. It is 5-megapixel, with focal ratio (the f-value is the ratio of the camera's focal length to the diameter of aperture opening) f/2.4, 28mm, autofocus, LED flash, 1/4' sensor size, geo-tagging, touch focus, powered by 1.2GHz dual-core Qualcomm Snapdragon S4 processor and it comes with 512MB of RAM, and packs 8GB of internal storage that can be expanded up to 64 GB via a micro SD card.

## 2.4. Dataset Description

Arabian horse is one of the oldest and purebred breeds of horses in the entire world dating back 3500 years [5]. It's originated on the Arabian Peninsula ( or Arabian island is a peninsula of

Western Asia located northeast of Africa on the Arabian region. The Arabian horse spread all over the world by war or trading. Due to speed, powerful, beauty, and strong bones of the Arabian horse many refinements are done to other breads by mating breads.

The Arabian lived in a desert climate and was prized by the nomadic Bedouin people so it has many distinguish characteristics. The head of the Arabian horse is extremely refined with wedge-shaped heads, a broad forehead, small ears, large eyes, large nostrils, and small muzzles. Most display a distinctive concave, or "dished" profile. The jibbah region lei a slight forehead bulge between their eyes adds additional sinus capacity and helps the Arabian horse to climate with its native dry desert home. The neck was arched which keep the large windpipe defined and clear to carry air to the lungs. This structure of the poll and throatlatch was called the mitbah or mitbeh by the Bedouin. The back was short with 17 pairs of ribs which provide plenty of room for lung expansion because of well sprung ribs and a deep chest cavity. The Arabian horse has greater density of bone than other breeds, short cannons, sound feet, long and level croup, height tail carriage and a broad, short back, all of which give the breed physical strength comparable to other horse breads. The standard height of the Arabian horse is standing from 14.1 to 15.1 hands (57 to 61 inches, 145 to 155 cm) [6]. The color of the Arabian horse varies between gray, white, sabino, Bay, and black. The standard weight is between 360 to 450 kilo grams as shown in figure4.

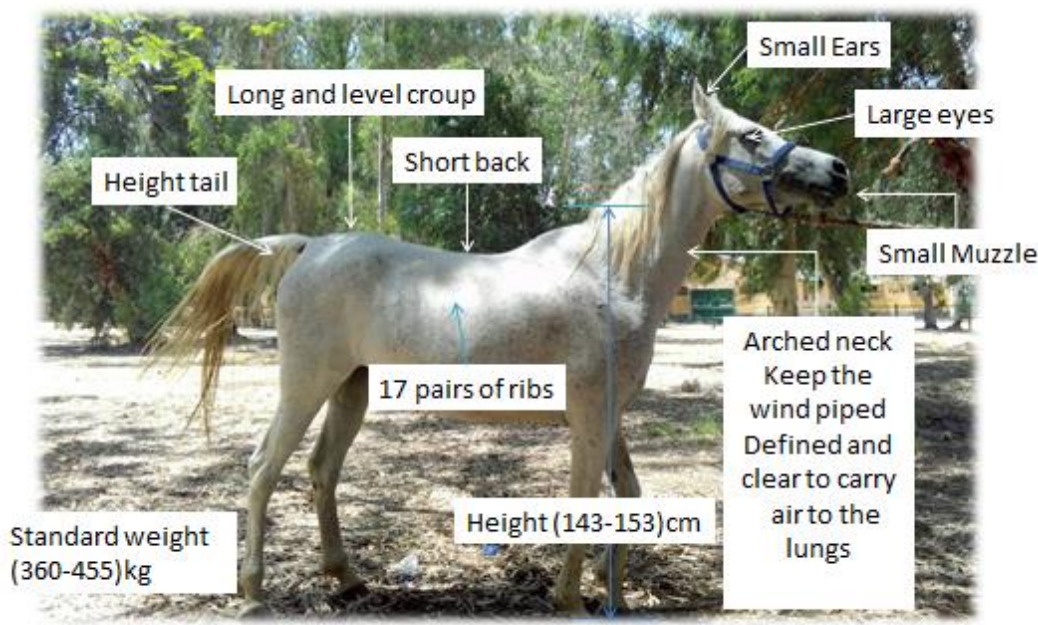

Fig.4. Arabian horse with distinguish characteristics.

The database has been collected from 50 (15 female and 35 male) Arabian horse with 6 live captured muzzle print images each. 300 Arabian horse muzzle print were imaged over a period of two weeks. The size of the images isn't the same, Class 1contain 15 female Arabian horse muzzle print images 6 images for each with minimum dimension is 600×700 pixels and maximum diminution is1400×1600 pixels. Class 2 also contains 35 male Arabian horse muzzle print images with minimum dimension 1000×1200 pixels and maximum dimension 1400×2100 pixels as shown in table 2. A special care has been given to the quality of the collected images. The collected images cover different quality levels and degradation factors such as image rotation and image partiality for simulating some real time identification conditions. Moreover, these images were performing some preprocessing operation such as, cropping, and given naming which works as FYXX, whereas XX is the cattle ID (1 to 15), and Y is the image order (1 to 6) for female horses, and MYXX for male horse whereas XX is the cattle ID (16 to 50), and Y is the image order (1 to 6).

**Conclusion**

Arabian horse muzzle print images have been collected from Al zahraa farm – Ain Shams – Egypt. Following three steps to get the images in good quality, first: make sure the horse stands steadfastly and raises its head. Second: wipe the nose before tokening image. Third: avoid infrared illumination and the distance between the camera and the horse was approximately 10 to 20 cm. By applying the three steps we get our datasets which consists of 300 colored images from 50 Arabian horses 15 female and 35 male with 6 live captured images each with different dimensions.